\documentclass{article}

\usepackage[final,nonatbib]{neurips_2023}
\usepackage[utf8]{inputenc}
\usepackage[T1]{fontenc}
\usepackage{hyperref}
\usepackage[all]{hypcap}
\usepackage{url}
\usepackage{booktabs}
\usepackage{amsfonts}
\usepackage{nicefrac}
\usepackage{microtype}
\usepackage{xcolor}
\usepackage{graphicx}

\title{Learning to Play Air Hockey with Model-Based \\ Deep Reinforcement Learning}

\author{
 Andrej Orsula \\
 Space Robotics Research Group (SpaceR) \\
 Interdisciplinary Centre for Security, Reliability and Trust (SnT) \\
 University of Luxembourg \\
  \texttt{andrej.orsula@uni.lu} \\
}

\hypersetup{
 pdfinfo ={
 Title={Learning to Play Air Hockey with Model-Based Deep Reinforcement Learning},
 Author={Andrej Orsula},
 Subject={Robot Air Hockey Challenge 2023},
 Keywords={Robotics, Reinforcement Learning},
 }
}

\begin{document}
\maketitle

\begin{abstract}
    In the context of addressing the Robot Air Hockey Challenge 2023, we investigate the applicability of model-based deep reinforcement learning to acquire a policy capable of autonomously playing air hockey. Our agents learn solely from sparse rewards while incorporating self-play to iteratively refine their behaviour over time. The robotic manipulator is interfaced using continuous high-level actions for position-based control in the Cartesian plane while having partial observability of the environment with stochastic transitions. We demonstrate that agents are prone to overfitting when trained solely against a single playstyle, highlighting the importance of self-play for generalization to novel strategies of unseen opponents. Furthermore, the impact of the imagination horizon is explored in the competitive setting of the highly dynamic game of air hockey, with longer horizons resulting in more stable learning and better overall performance. The source code and pre-trained models are available at~\href{https://github.com/AndrejOrsula/drl_air_hockey}{https://github.com/AndrejOrsula/drl\_air\_hockey}.
\end{abstract}

\section{Introduction}\label{sec:introduction}

The Robot Air Hockey Challenge 2023~\footnote{\url{https://air-hockey-challenge.robot-learning.net}} offers an engaging opportunity for robot learning researchers and practitioners to develop and evaluate their solutions for a realistic robot manipulation task. The challenge, organized by the Intelligent Autonomous Systems group at TU Darmstadt, consists of a virtual phase followed by real-world validation among the finalists. The organizers provide a simulation environment together with a baseline agent that can be used by the participants as a starting point for their own solutions. Furthermore, they facilitate the evaluation of sim-to-real transfer using a real-world robot air hockey platform. For each participating team, the challenge involves developing a controller capable of autonomously playing air hockey against controllers of other teams while employing a learning-based approach for at least one component of their agent.

There are several possibilities for combining various learning-based algorithms applied to distinct components of a performant controller. However, our initial interest revolves around techniques that directly optimize a single policy capable of playing the entire game of air hockey without relying on hand-crafted components or heuristics within the controller. From the popular approaches, the paradigms of imitation learning and reinforcement learning were initially considered. Despite the potential suitability of imitation learning, the need to collect expert demonstrations would necessitate additional engineering effort and could introduce undesired bias into the learned policy. Therefore, we decided to focus on reinforcement learning as it was deemed more suitable for the task of playing air hockey, which can be characterized as a two-player zero-sum game with simultaneously executed actions. This nature of the game further motivates the use of self-play that reinforcement learning agents can exploit to discover potential weaknesses in their own behaviour and develop better strategies to overcome them.

We believe that both model-free and model-based reinforcement learning approaches are applicable to the task of playing air hockey. In this work, we purposefully delimit our investigation to the model-based approach exemplified by DreamerV3~\cite{hafner2023mastering} due to recent advancements and its demonstrated capacity for solving various tasks in diverse domains. Although not directly analyzed in this work, model-based reinforcement learning might provide additional benefits in a competitive setting through its perceived sample efficiency that can also improve the adaptability of agents to the changing behaviour of their opponents.

\section{Challenge Setup}\label{sec:challenge-setup}

\begin{figure}[ht]
    \centering
    \includegraphics[width=1.0\linewidth]{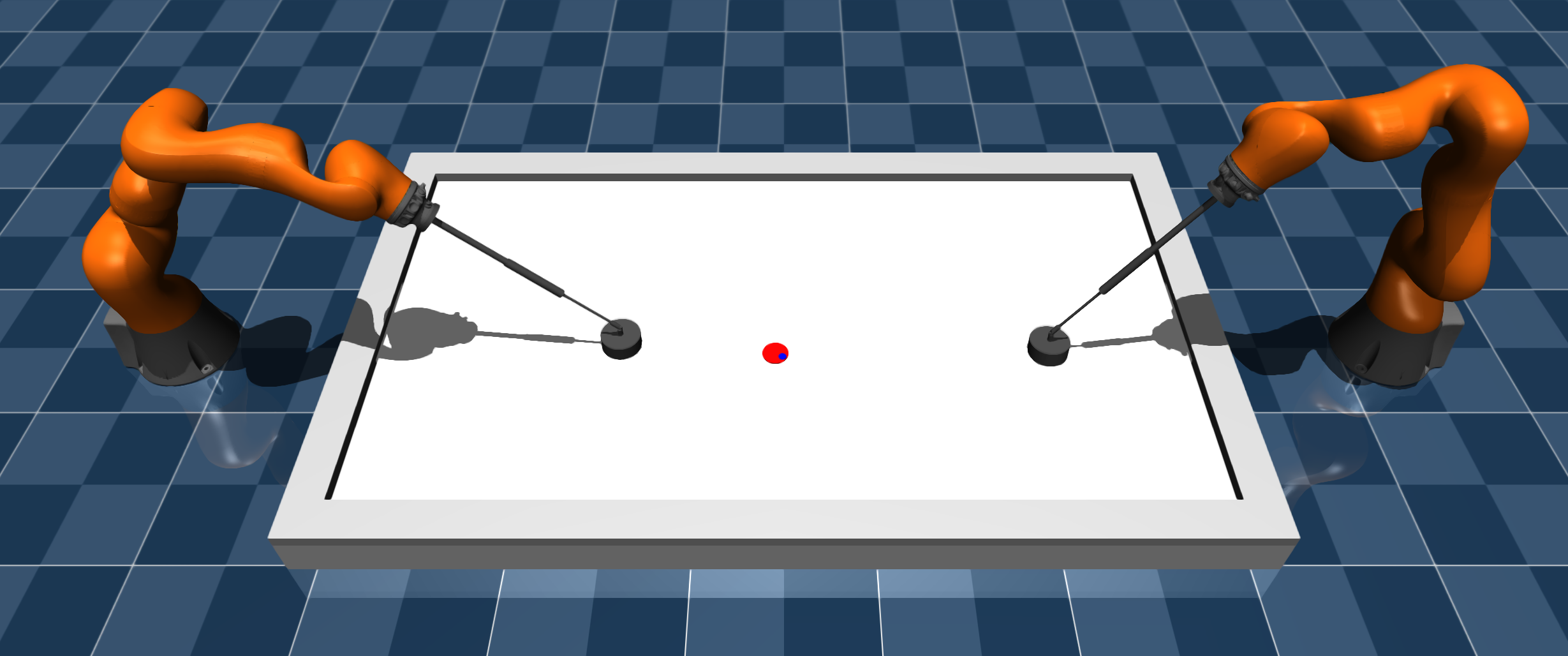}
    \caption{The simulation environment used in the Robot Air Hockey Challenge 2023 that features an air hockey table, puck, and two robotic manipulators controlled by the participating teams.}
    \label{fig:air_hockey_sim}
\end{figure}

The tournament phase of the challenge utilizes an air hockey table equipped with two \textit{KUKA iiwa14} \mbox{7-DoF} robotic manipulators, each controlled by one of the participating teams. This setup emulates the challenging nature of real air hockey with its fast-paced dynamics for robot-enabled gameplay. Each robot is equipped with a mallet attached to its end-effector that enables effective interaction with the puck. The simulation environment shown in Figure~\ref{fig:air_hockey_sim} is designed by the organizers to mirror the real-world setup inside \textit{MuJoCo}~\cite{todorov2012mujoco}, providing a virtual testbed for the development and evaluation of agents.

Both robots execute interpolated joint trajectories based on the commands supplied by the controllers of participants. The agents are required to provide joint-level commands at~\(50\)~Hz, imposing a computational time constraint of~\(20\)~ms per control cycle. At each step, the environment provides agents with the current state of the puck, the state of their joints, and the position of the opponent's mallet. The primary aim of agents is to score goals while preventing their opponent from doing so within the~\(15\)-minute duration of every match, which corresponds to~\(45000\)~simulation steps. Agents also receive a fault every time the puck remains on their side of the table for more than~\(15\)~s, with every three faults resulting in a subtracted point. Lastly, in cases where the puck gets stuck in the middle of the table while neither of the robots can reach it, the puck is randomly reset to one side.

In order to encourage the safety of sim-to-real deployment, the organizers impose limits on the behaviour of agents during the evaluation. These limits encompass the maximum joint positions and velocities, the Cartesian pose of the end-effector, and the aforementioned computational constraint. If agents exceed any of these limits, they accumulate penalty points and risk disqualification. Additionally, environmental stochasticity is introduced during the evaluation to examine the robustness of agents. Such modifications are undisclosed to participants but include environment disturbances, observation noise, and simulated loss of puck tracking.

\newpage

\section{Approach}\label{sec:approach}

Our approach revolves around leveraging DreamerV3 for training reinforcement learning agents to play the full game of robot air hockey. We focus on guiding agents towards optimal behaviour purely through sparse rewards corresponding to relevant score-affecting events that are not biased by potentially over-engineered reward shaping. To acquire policies capable of competing against various approaches, we employ a form of fictitious self-play to help agents discover more robust strategies by exploiting and correcting weaknesses in their prior behaviour.

The observability of the environment is facilitated for agents at each discrete time step through a set of low-dimensional environment states. Observation stacking is employed to provide additional temporal information about environment dynamics, even though it is not strictly necessary for the DreamerV3 agent that already benefits from the Markovian representation learned by its world model. Our agents interface with the robotic manipulator through continuous high-level actions for position-based control in the Cartesian plane. The inverse Jacobian method is then used to map these high-level actions into lower-level joint space commands that can be used to actuate the robot.

In the scope of this challenge, we train three distinct agents that optimize their behaviour towards different strategies. All of these agents use the same algorithm, network architecture, observations and action spaces but different reward functions that reinforce their behaviour towards a specific playstyle. We use the concept of multiple strategies in the context of self-play and within a multi-strategy ensemble described at the end of this section.

\subsection{Algorithm}\label{ssec:algorithm}

At its core, DreamerV3 consists of a world model, actor and critic networks that are optimized concurrently during the training. The world model learns a compact latent representation of the observations through autoencoding while enabling the prediction of future states and rewards via a sequence model with a recurrent state. Throughout the training, this model is used to generate abstract trajectories that are leveraged by the actor and critic to optimize their networks.

We believe that this process results in sample-efficient training that enables agents to learn how to compete in complex games and experience various environment transitions at a low computational cost, with the premise of acquiring a robust policy that generalizes well to novel strategies of unseen opponents. However, the soundness of this hypothesis and the full potential of this approach within the competitive setting of zero-sum games have not yet been thoroughly investigated. Therefore, we explore the benefits of model-based reinforcement learning in the context of robot air hockey.

\subsection{Observation Space}\label{ssec:observation-space}

The observation space of our agents comprises a set of low-dimensional environment states that the organizers made available for the participants. To capture information about the state of the controlled robot, its current joint positions and the planar position of the mallet attached as its end effector are included. Not only do these observations make the agent aware of its current state and manipulability, but it is also essential for the world model to learn the consequences of the agent's actions on the state of the controlled robot. The state of the opponent robot is available in the form of the planar position of their mallet, which allows agents to react to the opponent's actions even before they strike the puck. Similarly, the puck is observable through its planar position and orientation, providing the most crucial piece of information for agents to act upon and the world model to learn how to predict. Lastly, the time the puck spends on either side of the table is encoded in the observation space to provide agents with time awareness and help them avoid faults.

All observations are normalized to the range~\([-1, 1]\). The joint positions are normalized based on their limits, while the planar positions of the mallets and the puck are normalized based on the dimensions of the table. We encode the orientation of the puck as sine and cosine of the angle to preserve its continuity, while the duration of the puck spent on either side of the table is normalized based on the time limit until a fault would be accumulated with the sign corresponding to the side of the table.

Although the aforementioned states might provide a sufficient amount of information for agents to act upon, they lack temporal information about the environment dynamics required to fulfil the Markov assumption. Derivative states, such as the linear and angular velocity of the puck, are directly available to the participants and could also be estimated manually from two consecutive observations. However, their normalization would either require artificial limits or a more complex normalization scheme. Moreover, the stochasticity introduced during the evaluation in the form of simulated loss of puck tracking would make the derivative states unreliable. Therefore, we employ observation stacking~\cite{mnih2015human} as a simple method of providing additional temporal information about environment dynamics. In this way, the last~\(n\) observations of the selected states are stacked together and used as a single observation. At the beginning of each episode, the first observation is duplicated to fill the stack. Furthermore, as the position of the puck is considered especially important for agents while suffering from the loss of tracking, the observation stacking is performed asymmetrically with the aim of keeping an even longer history about this particular state. The final observation is then formed through the concatenation of all its components listed in Table~\ref{tab:observation-space}, resulting in a total dimensionality~of~\(40\).

\begin{table}[ht]
    \caption{The observation space of our agents, indicating the dimension and stack history of each state.}
    \label{tab:observation-space}
    \centering
    \begin{tabular}{lclc}
        \toprule
        \multicolumn{1}{c}{State}                 & \multicolumn{2}{c}{Dimension} & Stack (\(n\))                                   \\
        \midrule
        Participant --- Joint positions           & \hspace{1pt} 7 \hspace{-7pt}  & \hspace{-7pt} (\(DoF\))      & \hspace{1pt} 1   \\
        Participant --- Mallet position           & \hspace{1pt} 2 \hspace{-7pt}  & \hspace{-7pt} (\(x, y\))     & \hspace{1pt} 2   \\
        Opponent \hspace{1pt} --- Mallet position & \hspace{1pt} 2 \hspace{-7pt}  & \hspace{-7pt} (\(x, y\))     & \hspace{1pt} 2   \\
        Puck position                             & \hspace{1pt} 2 \hspace{-7pt}  & \hspace{-7pt} (\(x, y\))     & 10 \hspace{-1pt} \\
        Puck orientation                          & \hspace{1pt} 2 \hspace{-7pt}  & \hspace{-7pt} (\(sin, cos\)) & \hspace{1pt} 2   \\
        Fault timer                               & \hspace{1pt} 1 \hspace{-7pt}  & \hspace{-7pt} (\(t\))        & \hspace{1pt} 1   \\
        \bottomrule
    \end{tabular}
\end{table}

\subsection{Action Space}\label{ssec:action-space}

Playing air hockey incorporates continuous control over a planar surface. Therefore, we employ continuous high-level actions for position-based control in the Cartesian plane. At each discrete time step, agents are required to provide the absolute target position of their mallet as a 2D vector, with its coordinates normalized to the range~\([-1, 1]\) within the intersection of the table and the reachable workspace of the robot.

The high-level actions are then mapped into lower-level joint space commands using the inverse Jacobian method. Initially, the relative displacement of the mallet~\((\Delta x, \Delta y, \Delta z)\) is determined from its current position to the target. The position along the Z-axis is always constrained based on the relative height of the table. After computing the Jacobian matrix~\(J\) in the current joint configuration and obtaining its pseudo-inverse~\(J^+\), the target joint displacement can be calculated as~\((\Delta \theta_1, \Delta \theta_2, ..., \Delta \theta_7) = J^+ \cdot (\Delta x, \Delta y, \Delta z)\). To emphasize the displacement along the Z-axis and maintain consistent contact of the mallet with the table, a weighted product with weights~\((0.25, 0.25, 0.5)\) is applied. We ensure that the motion is dynamically feasible by clipping the joint displacements based on the joint velocity limits. Finally, the resulting joint positions and velocities are derived and used as low-level joint space commands, which are linearly interpolated over the duration of the control cycle.

\subsection{Reward Function}\label{ssec:reward-function}

To guide our agents towards optimal behaviour for playing air hockey, we adopt a sparse reward function designed to provide clear signals for specific events without introducing potential biases from intricate reward shaping. In the case of air hockey, the primary events coincide with terminal states that occur either when a goal is scored, or a fault is accumulated by one of the agents. Within the duration of a match, episode termination can also occur when the puck gets stuck in the middle of the table. However, we do not consider this terminal state as part of the reward. In general, our agents receive a positive reward for scoring the goal, while a negative reward is attributed for receiving a goal or causing a fault. Although a positive reward could also be attributed when the opponent receives a fault, we do not consider this event as it is dependent on the opponent's behaviour and could lead to ambiguous credit assignment.

We experiment with three different configurations of the reward function that reinforce agents towards different strategies, namely \textit{balanced}, \textit{aggressive} and \textit{defensive}. The \textit{balanced} strategy is used as the default configuration and is designed to encourage the agent to play a rounded game by scoring goals while defending their side of the table. Empirically, it was found that weighting the reward for scoring a goal equally with the reward for receiving a goal would result in risky behaviour where agents would keep their mallet close towards the centre of the table while eagerly trying to intercept the puck. This would often result in diminishing returns at the cost of receiving a goal, especially against novel opponents. Therefore, the \textit{balanced} strategy is incentivized to be slightly more defensive. On the other hand, the \textit{aggressive} strategy weights both rewards equally, resulting in riskier behaviour. Lastly, the \textit{defensive} strategy is designed solely to defend the goal against the opponent without any incentive to score goals. At the same time, all strategies receive negative rewards for causing a fault to discourage agents from accumulating them. The reward function is summarized in Table~\ref{tab:reward-function}.

\begin{table}[ht]
    \caption{The components of the reward function that reinforce agents towards different strategies.}
    \label{tab:reward-function}
    \centering
    \begin{tabular}{lccc}
        \toprule
        \multicolumn{1}{c}{Strategy} & Score a goal                     & Receive a goal & Cause a fault    \\
        \midrule
        Balanced (default)           & \hspace{-0.75pt}\(+\frac{2}{3}\) & \(-1\)         & \(-\frac{1}{3}\) \\[2pt]
        Aggressive                   & \hspace{-2.0pt}\(+1\)            & \(-1\)         & \(-\frac{1}{3}\) \\[2pt]
        Defensive                    & \(0\)                            & \(-1\)         & \(-\frac{1}{3}\) \\
        \bottomrule
    \end{tabular}
\end{table}

\subsection{Self-Play}\label{ssec:self-play}

Although a baseline agent is provided by the organizers, its policy covers only a limited subset of potential strategies in competitive robot air hockey. Training solely against this baseline agent would result in a policy effective only against this specific opponent, as demonstrated in our experimental section. To overcome this limitation, we employ a form of fictitious self-play~\cite{heinrich2015fictitious} to iteratively discover more robust strategies during training. In this way, our agents are trained against a pool of opponent agents with frozen weights. At the beginning of each episode, a new opponent is uniformly sampled from the pool and used until the episode terminates. The pool is gradually expanded with a new model every~\(1000\)~episodes while being limited to a total of~\(25\)~opponents, with a uniformly random opponent being replaced once the pool is full.

We also incorporate the strategies described in the previous section into self-play using a two-step procedure. First, three agents following the three strategies are trained using self-play, with the opponent pool initialized using the baseline agent. Once these agents begin to exhibit stable behaviour, their training is terminated while keeping a history of checkpoints for their models. Subsequently, a new agent is trained from scratch following the \textit{balanced} strategy. This time, the opponent pool is pre-filled with not only the baseline agent but also eight checkpoints for each strategy from the previously trained agents. In this way, the new agent is immediately exposed to a diverse pool of advanced opponents with expertise in different aspects of the game. The \textit{aggressive} opponents immediately incentivize the new agent to defend against their risky behaviour, while the \textit{defensive} opponents provide challenging agents to score goals against. At the same time, the \textit{balanced} opponents provide a tradeoff between the two extremes and further force the new agent to explore robust strategies.

Moreover, our custom version of the environment stochasticity is introduced during the training while incorporating environment disturbances, observation and action noise, and simulated loss of puck tracking. However, only the trained agent is affected by the added noise and simulated loss of puck tracking, while the opponents remain unaffected to further increase training difficulty.

\subsection{Multi-Strategy Ensemble}\label{ssec:multi-strategy-ensemble}

Recognizing the distinct strengths and weaknesses exhibited by agents trained with different strategies, we combine them into a single ensemble for the final evaluation of the challenge. This ensemble dynamically selects one of three playstyles, either \textit{balanced}, \textit{aggressive} or \textit{defensive}, based on the current score of the match estimated by the ensemble through the observations available to the participants. By default, the \textit{balanced} strategy is used for the majority of the game. If the opponent is in the lead, the ensemble switches to the \textit{aggressive} strategy to increase the likelihood of scoring a goal through a more risky playstyle. Conversely, if the opponent is trailing by a significant margin, the \textit{defensive} strategy is activated to provide a safer alternative in preventing the opponent from scoring a goal. With this design, the intended effect of the ensemble is to maximize the chances of winning by adapting the playstyle of the agent to the current state of the game.

\section{Experimental Evaluation}\label{sec:experimental-evaluation}

Throughout this section, we present the preliminary results of the proposed approach focused on the employed algorithm and selected aspects of the challenge.

\subsection{Training}\label{ssec:training}

Most default hyperparameters of DreamerV3 were used due to their demonstrated effectiveness. We primarily adjusted the size of the model to match the computational requirements of the challenge, resulting in a total of~\(9.7\)~M learnable parameters spread across the world model, actor and critic networks. With this model, the mean inference time of our agents was tested to be~\(5.2\)~ms on the CPU of the evaluation server across four matches against the baseline agent. We set the imagination horizon to~\(50\)~(\(1\)~s) in order to provide agents with a sufficient amount of time in which they can learn the direct consequences of their actions. Furthermore, we adjusted the training process by updating the replay capacity to~\(10^7\), the batch size to~\(32\), and the training ratio to~\(512\).

\begin{figure}[ht]
    \centering
    \includegraphics[width=0.9\linewidth]{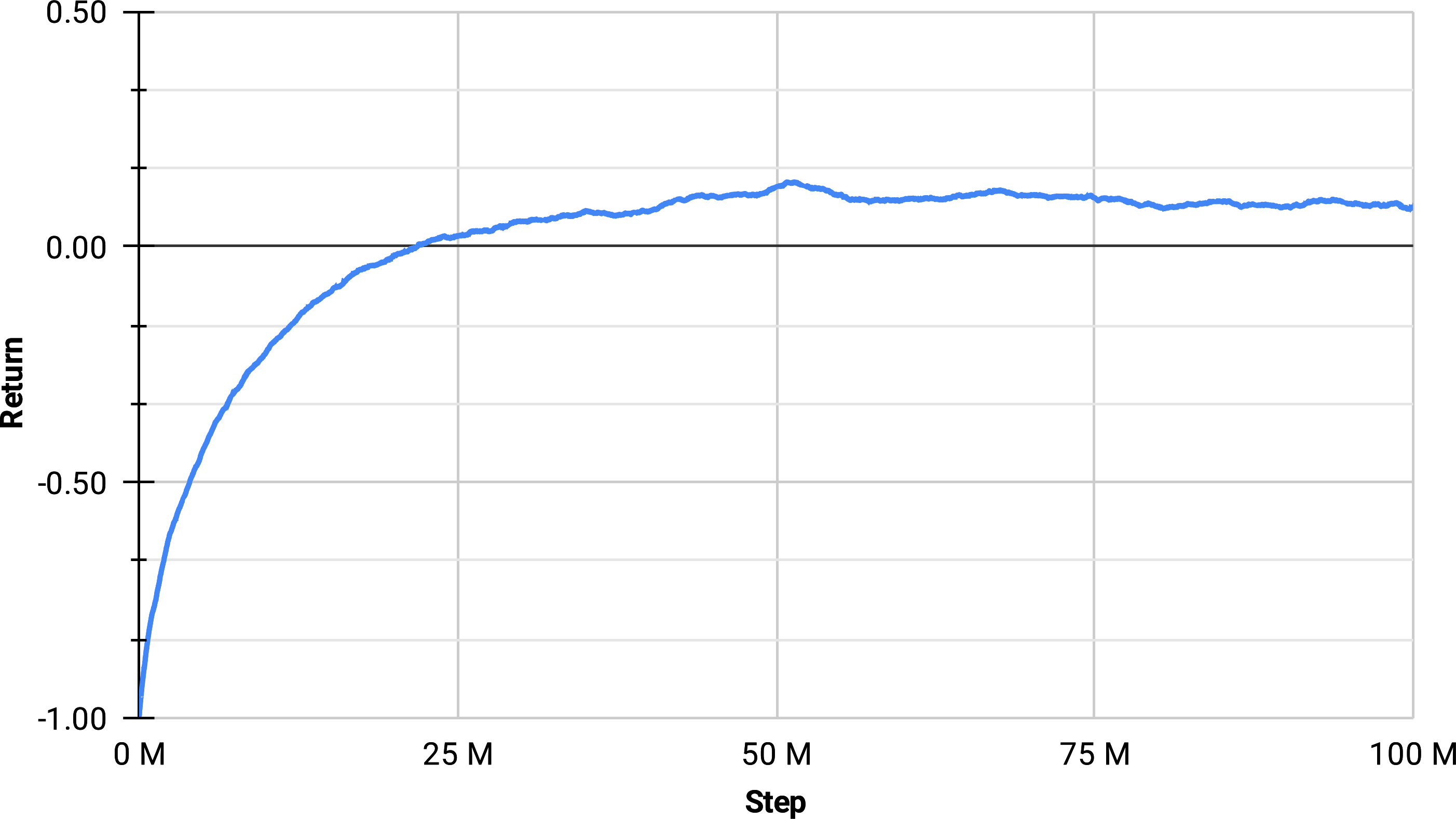}
    \caption{The learning curve of an agent following the \textit{balanced} strategy.}
    \label{fig:lc_balanced}
\end{figure}

Our agents are trained on a single workstation with \textit{AMD Ryzen 9 7950X} CPU and \textit{NVIDIA GeForce RTX 4090} GPU for a total of~\(100\)~M simulation steps. To accelerate the training and provide a diverse experience for agents, we employ ten parallel environment workers that collect transitions from the environment and store them in a shared replay buffer. Figure~\ref{fig:lc_balanced} contains the learning curve of the episodic reward achieved by an agent following the \textit{balanced} strategy throughout its training. It can be seen that the performance of the agent stagnates for most of its training, which can be largely attributed to the self-play mechanic. However, we note that the quality of the behaviour continues to improve. When tested against the baseline agent across~\(20\)~matches~(\(5\)~hours of gameplay), our agent wins with an average score of~\(6.1 : 0.2\).

\subsection{Effect of Self-Play}\label{ssec:effect-of-self-play}

\begin{figure}[ht]
    \centering
    \includegraphics[width=0.9\linewidth]{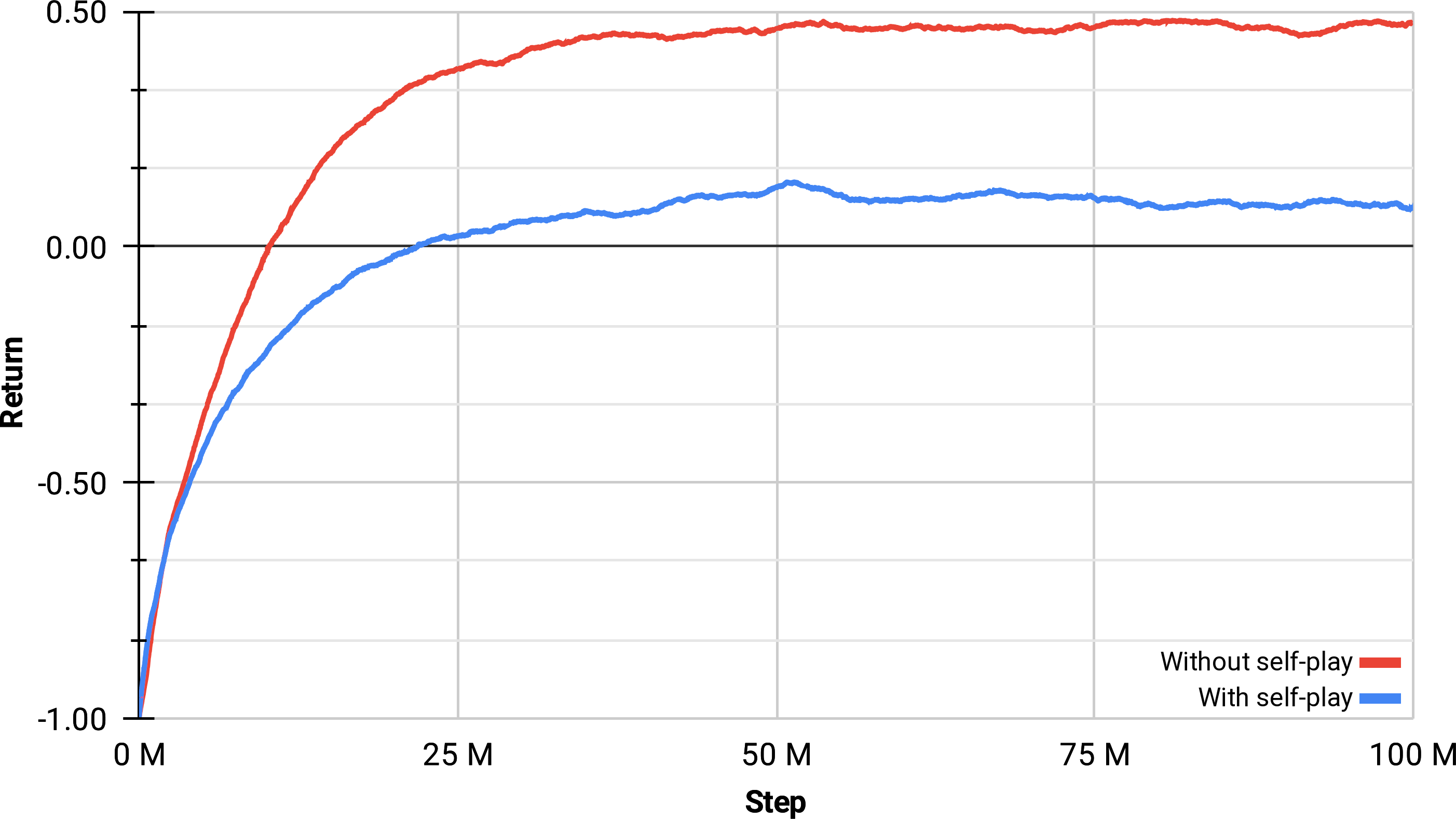}
    \caption{Learning curves of agents trained with and without self-play.}
    \label{fig:lc_selfplay}
\end{figure}

We also analyze the impact of self-play on the robustness of our agents. In addition to the agent from the previous section, we train another agent using the same configuration but without the self-play mechanism. This agent is trained only against the baseline agent for the entire duration of its training. When tested against the baseline agent across~\(20\)~matches, the agent without self-play wins with a much higher average score of~\(14.5 : 0.1\), which is indicative from its learning curve shown in Figure~\ref{fig:lc_selfplay}. However, when matched against the agent trained with self-play, it loses with an average score of~\(0.7 : 14.3\). This indicates that the agent trained without self-play is overfitted to the baseline agent and is unable to generalize to novel opponents, which disproves our initial hypothesis that learning purely from the randomized imagined sequences of the world model could enable agents to generalize to novel strategies of unseen opponents.

\subsection{Importance of Imagination Horizon}\label{ssec:importance-of-imagination-horizon}

\begin{figure}[ht]
    \centering
    \includegraphics[width=0.9\linewidth]{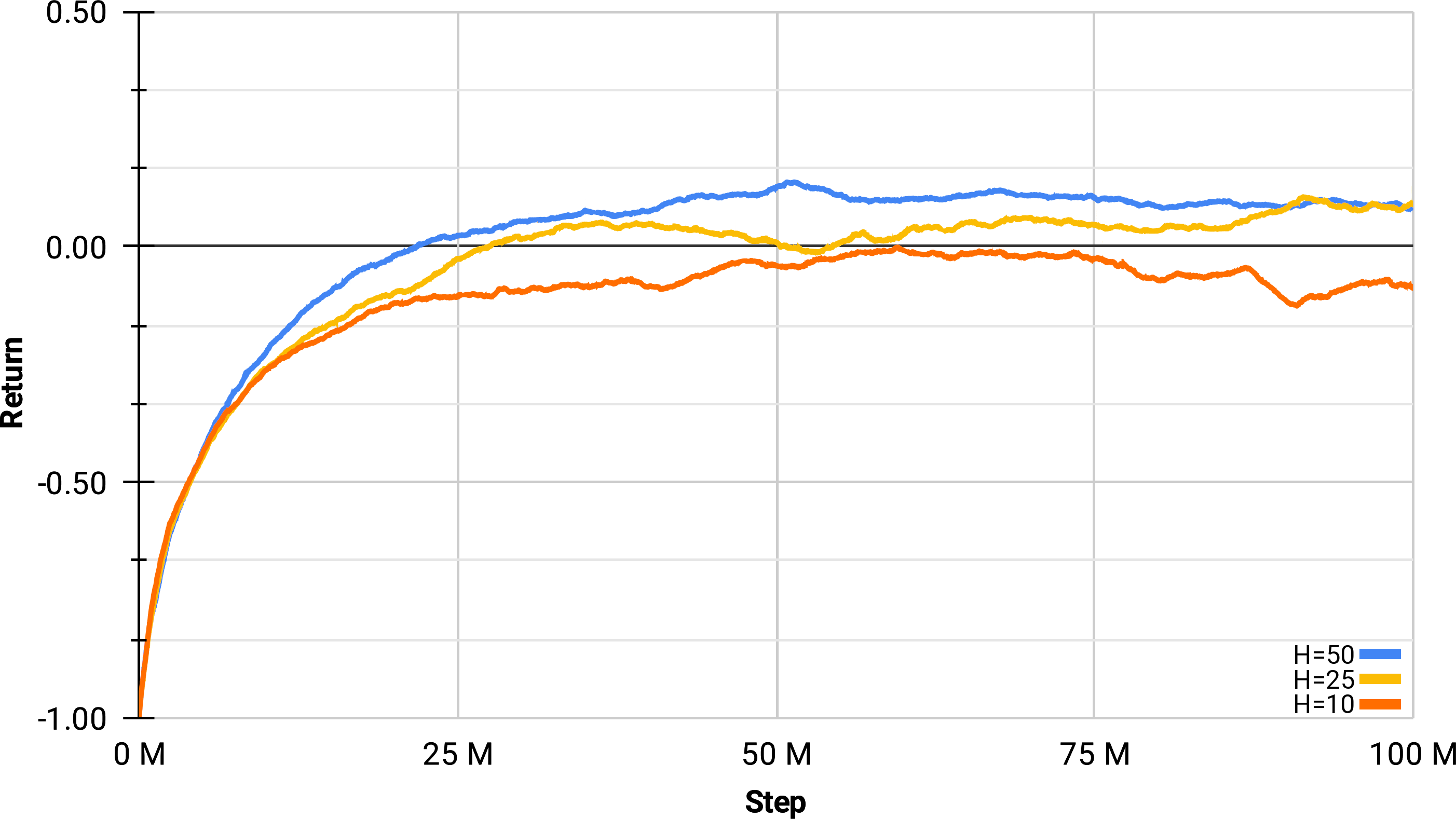}
    \caption{Learning curves of agents trained with different imagination horizon lengths (H).}
    \label{fig:lc_horizon}
\end{figure}

The critic network enables DreamerV3 to consider rewards even beyond the imagination horizon. However, the direct impact of this hyperparameter on the performance of the agent has not yet been explored to a great extent. Therefore, we investigate the importance of the imagination horizon in the context of robot air hockey. Figure~\ref{fig:lc_horizon} contains the learning curves of agents following the \textit{balanced} strategy trained with different lengths of the imagination horizon. It can be seen that agents trained with longer imagination horizons exhibit a more stable learning curve while also achieving a higher episodic reward. However, increasing the imagination horizon also increases the memory requirements of the agent during the training, where the length of~\(50\)~fully saturates the memory of the utilized GPU. When evaluated across~\(20\)~matches, the agent using the length of~\(50\) wins against agents using lengths of both~\(25\)~and~\(10\), with an average score of~\(3.9 : 2.0\) and~\(3.6 : 1.8\), respectively. Similarly, the agent using length of~\(25\)~marginally wins against the agent using length of~\(10\)~with a average score of~\(4.3 : 3.7\).

\newpage

\section{Results}\label{sec:results}

Within the tournament phase of the Robot Air Hockey Challenge 2023, following a double round-robin format, our agent achieved second place among seven participating teams that qualified for the final evaluation. Although we achieved the highest rank among purely learning-based approaches, our agent was repeatedly outperformed by teams with more performant controllers. Particularly, approaches leveraging optimal control and multiple policies in a hierarchical manner were found to be more effective compared to our single-policy approach.

The sim-to-real performance of our agent was also evaluated by the organizers on the real air hockey platform. In order to improve the smoothness of the executed trajectories, we incorporated an exponential moving average filter for the high-level actions of the agent. With this simple adjustment, the same agent was able to play air hockey on the real robot in a safe manner, albeit with a considerable delay in its responses. However, the performance gap of our agent between the simulated and real environment was significant, particularly due to the latency introduced by our filter and the inherent control cycle of the robot.

\section{Discussion and Conclusion}\label{sec:discussion-and-conclusion}

This work presented our approach employed in the Robot Air Hockey Challenge 2023. We demonstrated that model-based reinforcement learning can be applied to acquire a policy capable of competing in the challenging game of air hockey. Our findings indicate that agents are prone to overfitting when trained solely against a single playstyle, which indicates that self-play is essential for generalization to various strategies of potential opponents. Furthermore, the length of the imagination horizon was found to play a crucial role in the performance of agents.

The use of interpretable high-level actions simplifies the interface between the agent and the embodied system while providing a sufficient amount of control. However, using the policy to directly drive the motion of the robot results in a perceivably chaotic behaviour that raises safety concerns. Therefore, the incorporation of additional objectives or constraints into the policy and the change of the underlying low-level action interface should be explored.

Based on the evaluation and the achieved results of our agent, we believe that model-based reinforcement learning is a promising approach for solving robot manipulation tasks with complex dynamics and contact-rich interactions. The versatility of DreamerV3 for solving various tasks without the need for extensive hyperparameter tuning makes it a suitable candidate for applications in diverse domains. However, further evaluation is required to better understand its benefits within the context of competitive zero-sum games, especially with respect to the use of self-play and its eventual incorporation into the inner mechanism of learning from imagination.

\section*{Acknowledgements}\label{sec:acknowledgements}

We would like to thank the organizers of the Robot Air Hockey Challenge 2023 for providing an engaging platform for advancing the state-of-the-art in robot learning applied to robot manipulation.

\end{document}